\documentclass[letterpaper, 10 pt, conference]{ieeeconf}
\IEEEoverridecommandlockouts                              
\usepackage[utf8]{inputenc} 
\usepackage[T1]{fontenc}    
\usepackage{url}            
\usepackage{booktabs}       
\usepackage{amsfonts}       

\usepackage{graphicx}
\usepackage{amsmath}
\usepackage{amssymb}
\usepackage{xcolor}
\usepackage{accents}
\usepackage{caption}
\usepackage{xcolor}
\usepackage{comment}
\usepackage{multirow}
\usepackage{subcaption}
\usepackage{soul}
\usepackage{color}
\usepackage{leftidx}

\usepackage{threeparttable}
\usepackage{makecell}
\usepackage{nicefrac}
\usepackage[normalem]{ulem}

\newcommand\mb{\mathbf}

\title{\LARGE \bf Double-Iterative Gaussian Process Regression for Modeling Error Compensation in Autonomous Racing}
\author{
Shaoshu Su$^1$, Ce Hao$^2$, Catherine Weaver$^2$, Chen Tang$^2$, Wei Zhan$^2$, Masayoshi Tomizuka$^2$%
\thanks{$^1$ S. Shu is with the Department of Electrical and Computer Engineering, University of California, Riverside, CA 92521 USA {\tt\footnotesize \{shaoshu.su@email.ucr.edu\}}.}%
\thanks{$^2$ C. Hao, C. Weaver, C. Tang, W. Zhan, M. Tomizuka are with the Department of Mechanical Engineering, University of California Berkeley, CA, USA {\tt\footnotesize \{cehao, catherine22, chen\_tang, wzhan, tomizuka\}@berkeley.edu.}}%
\thanks{* Work conducted during Shaoshu Su's visit at UC Berkeley.}%
\thanks{** Supported by Sony AI, Sony Research Inc.}%
}

\begin{document}
\maketitle

  

\begin{abstract}                
Autonomous racing control is a challenging research problem as vehicles are pushed to their limits of handling to achieve an optimal lap time; therefore, vehicles exhibit highly nonlinear and complex dynamics. Difficult-to-model effects, such as drifting, aerodynamics, chassis weight transfer, and suspension can lead to infeasible and suboptimal trajectories. While offline planning allows optimizing a full reference trajectory for the minimum lap time objective, such modeling discrepancies are particularly detrimental when using offline planning, as planning model errors compound with controller modeling errors. Gaussian Process Regression (GPR) can compensate for modeling errors. However, previous works primarily focus on modeling error in real-time control without consideration for how the model used in offline planning can affect the overall performance. In this work, we propose a \textit{double-GPR} error compensation algorithm to reduce model uncertainties; specifically, we compensate both the planner's model and controller's model with two respective GPR-based error compensation functions. Furthermore, we design an iterative framework to re-collect error-rich data using the racing control system. We test our method in the high-fidelity racing simulator Gran Turismo Sport (GTS); we find that our \textit{iterative, double-GPR} compensation functions improve racing performance and iteration stability in comparison to a single compensation function applied merely for real-time control. 
\end{abstract}

\section{Introduction}\label{Sec:sec1} 
Recently, autonomous vehicle racing has drawn increased attention as a complex control problem \cite{betz2022autonomous}. Often, the problem of time-trial racing is explored, where the goal is to control a single race car around a constrained track in the fastest time. Compared to autonomous driving in urban or highway environments, such time-trial racing requires pushing the vehicle to the limits of handling in order to rapidly traverse the course of a racetrack. Researchers have shown that model-free reinforcement learning approaches are able to develop autonomous policies by iteratively improving their policies that can outperform even the top human racers in both time-trial and multi-opponent races \cite{wurman2022outracing}. Yet such end-to-end approaches require potentially dangerous exploration and many trials to collect sufficient data, and it can be difficult to interpret the most important factors of vehicle dynamics and control from resulting neural network policy. Therefore, many researchers are exploring how to bridge the possible gap between structured model-based planning and control and end-to-end learning \cite{betz2022autonomous}. 

One crucial challenge in model-based racing lies in identifying an accurate yet simple vehicle dynamics model. Racing vehicles operate in highly nonlinear regions of vehicle dynamics, where complex factors such as drifting, aerodynamics, chassis weight transfer, and suspension have a non-negligible effect. However, it is typically infeasible to solve an optimization problem for planning or control using a descriptive model including all components of a race car~\cite{milliken1995race}. Simplified models, for instance a single track bicycle model \cite{rajamani2011vehicle} and empirical tire friction models such as Pacejka's Magic Formula \cite{pacejka2005tire} aid researchers in developing models that are feasible for planning and control of racing vehicles \cite{betz2022autonomous}. However, such simplifications can fail to model the most extreme dynamics often occurring at the most critical corners of a racetrack, which can lead to suboptimal performance \cite{ourpriorwork}. 

\begin{figure*}[t]
    \centering
    \includegraphics[width=0.72\textwidth]{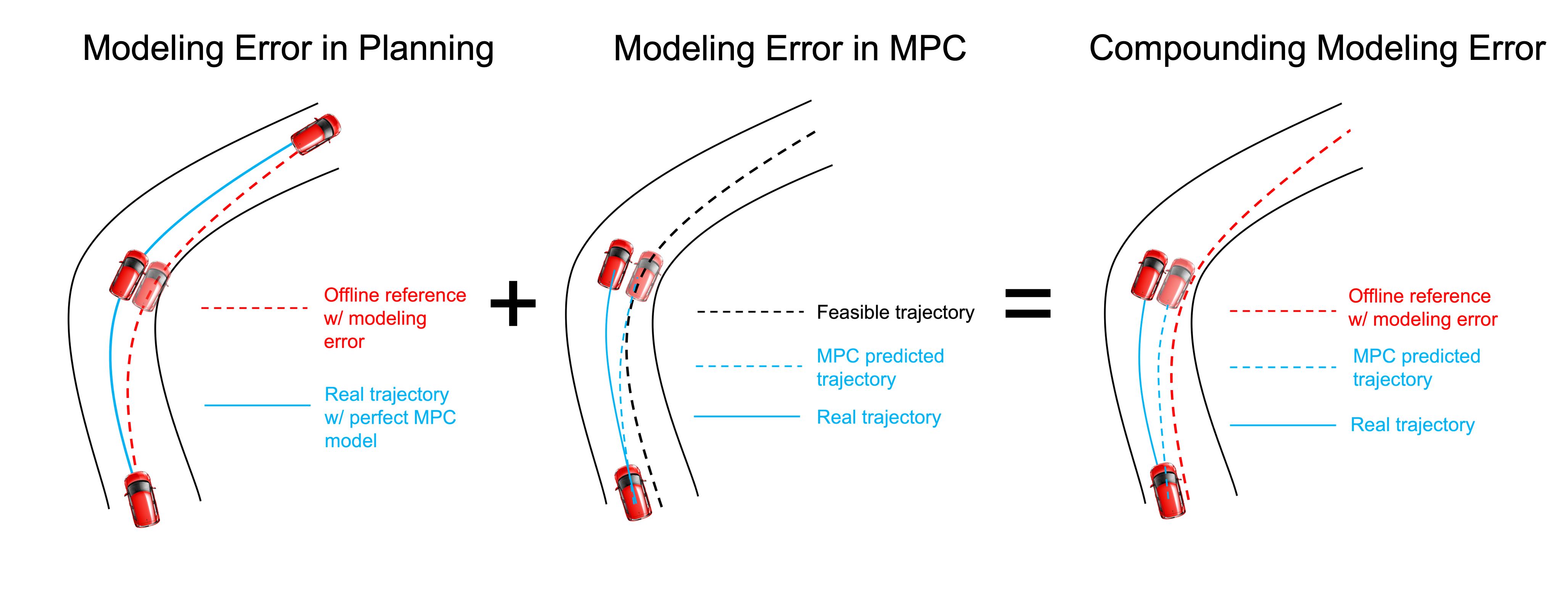}
    \caption{Example of compounding planning and control modeling errors. \textbf{Left}: modeling errors in the offline planner may lead to an infeasible reference trajectory, which cannot be tracked well even by an MPC controller using a perfect oracle of the vehicle dynamics; \textbf{Middle}: modeling errors in the MPC controller cause incorrect prediction of the vehicle state, leading to tracking errors even when the reference is dynamically feasible; \textbf{Right}: The modeling errors in the planning and control phases compound, undermining the overall racing performance.}
    \label{Fig:concept_figure}
\end{figure*}

Furthermore, while some researchers have developed control objectives that tend to perform well on the racing objective, for example by encouraging course progress over a relatively short control horizon \cite{hewing2018cautious}, offline trajectory planning of the full racing trajectory allows for directly optimizing of the lap time, generating a time-optimal reference trajectory over the entire racetrack. In contrast to tracking the centerline, using an online controller to track the time-optimal reference trajectory leads to competitive racing performance \cite{ourpriorwork}. However, separating the racing problem into an offline optimal planning module and a real-time tracking module leads to compounding modeling errors due to the reliance on the model by both the planner and controller (Fig.~\ref{Fig:concept_figure}). The compounding effect further exaggerates the impact of modeling discrepancy on racing performance.

One promising solution to tackle this problem is introducing a modeling-error compensation function in addition to the nominal descriptive dynamics model.
Recently, Gaussian Process Regression (GPR) is commonly used to learn a modeling-error compensation function.
One compelling advantage of GPR model is it allows convenient online update when new observations arrive, which enables iterative improvement of modeling accuracy and control performance through online learning~\cite{kabzan2019learning}.
However, such works primarily focus on modeling error in real-time control, without consideration for how the model of an offline optimal trajectory planner can affect the overall control performance. As mentioned above, it is equally important to use an accurate vehicle model in the offline planning stage at the first place, due to the compounding effect of modeling errors. Therefore, we focus our efforts reducing modeling errors with GPR, with special consideration that models should be reliable for both planning and control. 

In this work, we present GPR-based error compensation notably with two components: 1) error compensation for the low-level real-time controller, and importantly 2) error compensation for the \textit{offline}, optimal trajectory planner. We term such two-part compensation \textit{double GPR}. Furthermore, we find that data collection that is representative of the dynamics at the critical dynamic regions is essential for adequate compensation of modeling errors in those regions; therefore, we present an iterative framework that can guide data collection to improve the modeling for both planning and control, which we call \textit{iterative, double GPR}. We incorporate the proposed error compensation modules into the racing planning and control framework proposed in our prior works~\cite{ourpriorwork}, and test it in the highly realistic racing simulation platform, Gran Turismo Sport~\cite{GTS_web}, where we show that \textit{iterative, double GPR} yields lower lap times and reduced modeling errors in comparison to single or one-shot compensation on either the planning or control modules.

%
\section{Related works}\label{Sec:RelatedWorks}

Learning-based dynamics models have been adopted in robotics applications for a long time~\cite{kappler2017new,li2019model,rastogi2018sample,ha2015reducing}. 
Among them, Gaussian Process Regression (GPR) is commonly adopted for complex robotics systems
\cite{lima2020sliding,jing2016calibration,fang2019vision,wan2017optimal}.
In the field of autonomous driving, the majority of prior works focus on learning GPR-based modeling error compensation for online control. \cite{hewing2018cautious} proposes a stochastic model predictive control (MPC) algorithm with a GPR-based residual model, where a chance constraint is formulated to take the modeling uncertainty estimated in GPR into consideration to improve safety. The proposed method is further extended and validated on a full-size driverless race car~\cite{kabzan2019learning}. To make GPR feasible for real-time computation, the GPR-based residual model is implemented based on sparse GPs~\cite{pmlr-v5-titsias09a} and learnt with a small dataset. The controller in~\cite{kabzan2019learning} is designed based on model predictive contouring control, which simultaneously plans a trajectory to encourage course progress over the control horizon and implements control in a receding fashion. However, the relatively short horizon can lead to suboptimal performance, as future track geometry is not considered outside the control horizon. In \cite{jain2020bayesrace}, GPR-based compensation function is adopted to a so-called extended kinematic vehicle model; here the modeling compensation is also applied only to short-horizon MPC, and the racing line is planned offline and set prior to real-time control. Therefore, modeling errors present during offline planning may lead to an infeasible or suboptimal trajectory. 

Conversely, some researchers explore the trajectory planning problem with GPs. \cite{wischnewski2019model} uses GP to directly model the acceleration limits of a scaled RC car; the minimum lap time problem is approximated by using a factor of the accelerating limits, yielding an online trajectory planner that can lead to vehicle stability and quality tracking performance. However, the approximated objective does not necessarily minimize the true lap time, and the oversimplification of the dynamic bicycle model into an extended kinetic model does not account for force that have shown to be highly influential in planning and control, such as aerodynamic force and longitudinal weight transfer \cite{ourpriorwork}.

\section{Background}\label{Sec:background} 
In this section, we introduce Gaussian Process Regression (GPR) for error compensation and describe the kinetics and dynamics of racing vehicle model.
\subsection{Gaussian Process Regression}
Gaussian Process Regression (GPR) is a non-parametric learning method. We only give a concise introduction here primarily to help the readers get prepared for the technical content introduced in the following sections. The readers could refer to \cite{williams2006gaussian} and \cite{wang2020intuitive} for more details. 

Given a feature vector $\mb{z} \in \mathbb{R}^{n_f}$ with ${n_f}$ being its number of dimensions, and a output vector $\mb{y} \in \mathbb{R}^{n_d}$ with $n_d$ being its number of dimension, we assume they are related as follows:
\begin{equation*}
    \mb{y} = \mb{g}(\mb{z}) + \boldsymbol{\omega}
\end{equation*}
where $\mb{g}(\mb{z})$ is an unknown stochastic function and $\boldsymbol{\omega}\in\mathbb{R}^{n_d}$ is i.i.d. Gaussian noise with zero mean and diagonal covariance $\Sigma_w=\{\sigma_1^2 \dots \sigma_{n_d}^2\}$. In Gaussian Process Regression, we parameterize the function $\mb{g}(\mb{z})$ with a Gaussian distribution as 
\begin{equation}
\begin{aligned}
\mb{g}(\mb{z})\sim 
\mathcal{N}(\boldsymbol{\mu}(\mb{z}), \mb{\Sigma}(\mb{z})),
\end{aligned}
\end{equation}
where $\boldsymbol{\mu}(\mb{z}) = [\mu^1(\mb{z}),\dots,\mu^{n_d}(\mb{z})  ] \in \mathbb{R}^{n_d}$ 
and $\mb{\Sigma}(\mb{z}) = \text{diag}([\Sigma^1(\mb{z}),\dots,\Sigma^{n_d}(\mb{z})]) \in \mathbb{R}^{{n_d}\times {n_d}}$.

Given a finite dataset $\mathcal{D}$ of size $m$ consisting of feature-output tuples, $ \{(\mb{z}_1, \mb{y}_1), \dots , (\mb{z}_m, \mb{y}_m)\}$, we denote it as $\mathcal{D}=\{\mb{Z}, \mb{Y}\}$ with input features $\mb{Z}= [\mb{z}^\top_1; \dots; \mb{z}^\top_m ] \in \mathbb{R}^{m \times n_f}$, and output data $\mb{Y}= [\mb{y}^\top_1; \dots; \mb{y}^\top_m ] \in \mathbb{R}^{m \times n_d}$. GPR use the dataset $\mathcal{D}$ to fit $\boldsymbol{\mu}(\mb{z})$ and $\mb{\Sigma}(\mb{z})$ as
\begin{align}
\mu^a(\mb{z}) &= \mb{k}^{a}_{\mb{z}\mb{Z}}(\mb{K}^{a}_{\mb{Z}\mb{Z}} + \mb{I} \sigma^2_a )^{-1}[\mb{Y}]_{.,a},\\
\Sigma^a(\mb{z}) &= k^{a}_{\mb{z}\mb{z}}-\mb{k}^{a}_{\mb{z}\mb{Z}}(\mb{K}^{a}_{\mb{Z}\mb{Z}} + \mb{I} \sigma^2_a)^{-1}\mb{k}^{a}_{\mb{Z}\mb{z}},
\end{align}
for $a=1,...,n_d$. In the equations above, $[\mb{Y}]_{.,a}$ is the $a$-th column of $\mb{Y}$, and $\mb{K}^{a}_{\mb{Z}\mb{Z}}$ is the Gram matrix. For the element at the $i$-th row and $j$-th column of the Gram matrix, we have $[\mb{K}^{a}_{\mb{Z}\mb{Z}}]_{ij} = k^a(\mb{z}_i,\mb{z}_j)$, where $k^a(\cdot, \cdot)$ is a kernel function. Likewise, the vector $\mb{k}^j_{\mb{Z}\mb{z}}=(\mb{k}^j_{\mb{z}\mb{Z}})^\intercal\in\mathbb{R}^m$ has its $j$-th element defined as $[\mb{k}^{a}_{\mb{Z}\mb{z}}]_{j} = k^a(\mb{z},\mb{z}_j)$, and we define $k^{a}_{\mb{z}\mb{z}} =  k^a(\mb{z},\mb{z}) \in \mathbb{R}$.

Regarding the kernel function, there are multiple candidates in the literature. The choice of kernel depends on the specific problem and assumed distribution of the collected data. In this work, we adopt the most widely used kernel, the Radial Basis Function (RBF):
\begin{equation}
\begin{aligned}\label{equ:kernel_func}
k^a(\mb{z}_i,\mb{z}_j) = \sigma^2_{ka} \text{exp}(-\frac{1}{2}(\mb{z}_i-\mb{z}_j)^\top\mb{L}^a (\mb{z}_i-\mb{z}_j)),
\end{aligned}
\end{equation}
where $\mb{L}^a$ is a diagonal matrix with positive length scale parameters as its diagonal elements and $\sigma^2_{ka}$ is the squared signal variance. The values of priors $\sigma_{a}$ and $\sigma_{ka}$, and scaling factor $\mb{L}^a$ in the kernel function \eqref{equ:kernel_func} are obtained by maximizing a marginal likelihood function as in \cite{williams2006gaussian}: 
\begin{equation}\label{para_optm_func}
\text{log}\;p(\mb{Y}^a|\mb{Z},\boldsymbol{\theta}) = 
-\frac{1}{2}{\mb{Y}^a}^\top 
\mb{K}^{-1} \mb{Y}^a -\frac{1}{2} \text{log} \; |\mb{K}| - \frac{n}{2}\text{log}2\pi,
\end{equation}
where $\boldsymbol{\theta}=\{ \sigma_{a}$,  $\mb{L}^a$, $\sigma_{ka}\}$ and $\mb{K} = \mb{K}^{a}_{\mb{Z}\mb{Z}} + \mb{I} \sigma^2_a$.

\subsection{Racing Vehicle Model}
\label{Sec:VehicleModel}
In this section, we introduce a nominal vehicle model that is used as the foundation of the model-based planning and control in the following Section \ref{Sec:TrajPlan} and \ref{Sec:Controller}. 

\begin{figure}[th!]
    \centering
    \includegraphics[width = 0.45\textwidth]{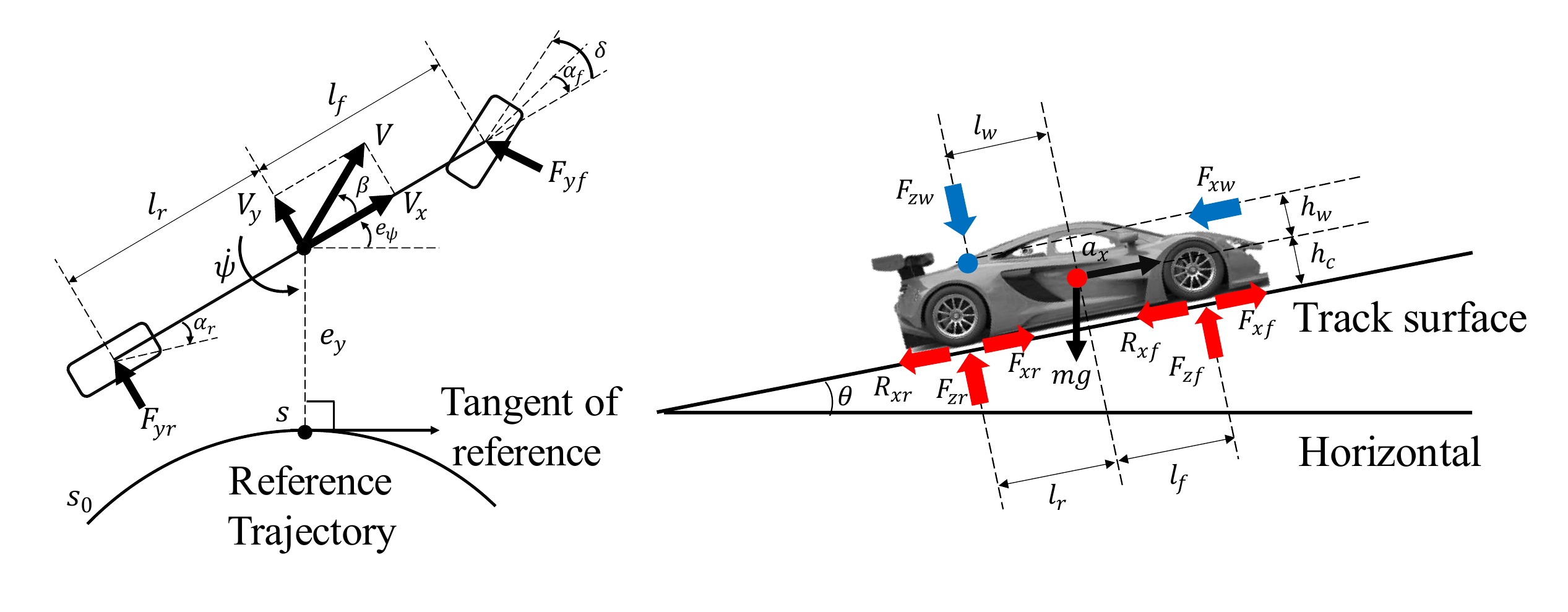} 
    \caption{Diagram of racing vehicle model. \textbf{Left}: Overview and relative position to the reference trajectory; \textbf{Right}: Lateral profile with track surface pitch angle.}
    \label{fig:vehicle model}
\end{figure}

The vehicle bicycle model assumed a single-track vehicle state as shown in Fig.~\ref{fig:vehicle model}. 
In particular, we model a vehicle that is front-wheel steered and rear-wheel driven, although the model can easily be extended for other drivetrains. The vehicle states and control variables are
\begin{equation*}
    \mb{x} = [V_x, V_y, \dot{\psi}, e_\psi, e_y, s]^\intercal, \quad
    \mb{u} = [\delta, {a}_x]^\intercal.
\end{equation*}
The state $\mb{x}$ consists of the longitudinal and lateral velocities, $V_x$ and $V_y$ respectively, in the vehicle's body frame, the yaw angular velocity $\dot{\psi}$, and the relative yaw angle and distance from the vehicle to a reference point ${e}_{\psi}, {e}_{y},$. The control input $\mb{u}$ consists of the steering angle $\dot{\delta}$ and the longitudinal acceleration $a_x$.

The longitudinal and lateral dynamics are derived from force-mass and inertia-moment balance and are given by 
\begin{subequations}
\begin{align}
\dot{V}_{x}=&a_x - \frac{1}{m}\left(F_{y f} \sin (\delta)+R_{x}+F_{x w}\right) \nonumber \\ 
&-g \sin (\theta)+\dot{\psi} V_{y}, \label{Equ:long_model}\\ 
\dot{V}_{y}=&\frac{1}{m}\left(F_{y f} \cos (\delta)+F_{y r}\right)-\dot{\psi} V_{x}, \label{Equ:lat_model1} \\
\ddot{\psi}=&\frac{1}{I_{z z}}\left(l_{f}F_{y f} \cos (\delta)-l_{r} F_{y r}\right). \label{Equ:lat_model2}
\end{align}
\end{subequations}
Parameters specific to the vehicle include mass $m$, moment of inertia $I_{zz}$, and the distance to the center of mass (CoG) to the front axle, $l_f$, and rear axle, $l_r$. The forces acting on the vehicle include the 
 the tire rolling resistance, $R_x$, the wind drag force on the vehicle body, $F_{xw}$, and gravity that acts with acceleration $g$.
The lateral tire forces of the front and rear tires, $F_{yf}$ and $F_{yr}$, are nonlinear and vary as the tire slips along the road surface.
To learn more details about the nonlinear model used to describe these forces, readers could refer to our earlier work \cite{ourpriorwork}.

From the velocities in the body frame ($V_x$ and $V_y$) and the yaw rate ($\dot{\psi}$), the positional motion of the vehicle is derived from kinematics. In particular, we similarly adopt the Frenet coordinate to model vehicle position with respect to a reference path, as depicted in Fig. \ref{fig:vehicle model}. 
The relative yaw angle and distance, $e_{\psi}$ and $e_y$ respectively, of the vehicle are determined with respect to the nearest point (to the CoG of the vehicle) on the reference as defined in~\cite{ourpriorwork}. The derivatives of these states are derived from kinematics as 
\begin{align}
\dot{e}_{\psi}&=\dot{\psi}-\frac{V_{x} \cos \left(e_{\psi}\right)-V_{y} \sin \left(e_{\psi}\right)}{1-\kappa(s) e_{y}} \kappa(s) \label{Equ:epsi} \\
\dot{e}_{y}&=V_{x} \sin \left(e_{\psi}\right)+V_{y} \cos \left(e_{\psi}\right) \label{Equ:ey}
\end{align} 
where $s$ denotes the vehicle's progress along the reference path from the starting point, and it is modeled as
\begin{equation}
     \dot{s}=\frac{V_{x} \cos \left(e_{\psi}\right)-V_{y} \sin \left(e_{\psi}\right)}{1-\kappa(s) e_{y}}, \label{Equ:s}
\end{equation}
and $\kappa$ denotes the curvature at the reference point 
at $s$. Equations ~\eqref{Equ:long_model} through \eqref{Equ:s} are summarized as
\begin{equation}\label{equ:model_summary}
     \dot{\mb{x}}=f(\mb{x}, \mb{u}).
\end{equation}
In practice, we further discretize the model for the purpose of planning and control as the following
\begin{equation}\
\quad \mb{x}_{k+1} = \mb{x}_{k} + f(\mb{x}_{k},\mb{u}_{k}) \Delta t.
\\
\end{equation}

\section{Learning-based Trajectory Planning and Control Algorithms}\label{Sec:Plan_MPC}

In this section, we detail the planning and control modules that exploit our \textit{iterative, double GPR} method for autonomous racing. We employ GPR to provide two error compensation models; we then use these models in two modules: a 1) time-optimal trajectory planner and a 2) reference-tracking model predictive controller (MPC). In the following subsections, we first detail the planning and control frameworks, then describe the \textit{double-GPR} based modeling error compensation for both planning and control, and then describe the \textit{iterative} process for adapting the compensation models.

\subsection{Time-Optimal Trajectory Planning}\label{Sec:TrajPlan}
We aim to plan a trajectory by directly solving a motion planning optimization problem to minimize the lap time, $T$, which can yield competitive racing results \cite{ourpriorwork}. Similarly, we adopt a close approximation of the minimum lap time objective that is derived by transforming the integral over the trajectory's horizon in \textit{time} to an integral in the \textit{spatial domain}~\cite{kapania2016sequential, ourpriorwork}. We likewise find a feasible initial guess by minimizing the curvature of a racing path and determining a feasible velocity for the path to reduce the potential for falling into suboptimal local optima. We summarize the minimum time objective here:
\begin{subequations}
\begin{align}
    \min_{\{\mb{x}_i\}_{i=1}^N,\{\mb{u}_{i}\}_{i=1}^N} \sum_{k=1}^{N} & \frac{\left(1-\kappa_k \ e_{y, k}\right) \Delta s_k}{V_{x, k} \cos \left(e_{\psi, k}\right)-V_{y, k} \sin \left(e_{\psi, k}\right)} \label{Equ:TimeObj} \\
    \mathrm{ s.t. } \quad \mb{x}_{k+1} & = \mb{x}_{k} + f(\mb{x}_{k},\mb{u}_{k}) \Delta t, \label{Equ:TimeObj_constr1}\\
    \mb{x}_{1} & =\mb{x}_{N}, \label{Equ:TimeObj_constr2}\\
    w_{r,k} & \leq e_{y,k} \leq w_{l,k}, \label{Equ:TimeObj_constr3}\\
    \delta_{\min,k } & \leq \delta_k \leq \delta_{\max,k }. \label{Equ:TimeObj_constr4}
\end{align}
\end{subequations}

In this motion planning problem, the constraints \eqref{Equ:TimeObj_constr2}, \eqref{Equ:TimeObj_constr3}, and \eqref{Equ:TimeObj_constr4} respectively ensure the state starts and ends at the same point, the vehicle remains within the track width on the left $w_{l,k}$ and right $w_{r,k}$, and the steering angle remains in the allowable range. The equality constraint \eqref{Equ:TimeObj_constr1} enforces that the states obey the vehicle model (\ref{equ:model_summary}). Since the time-optimal trajectory is constrained to remain feasible to the model, an accurate model is critical to result in a trajectory that is not only feasible, but is actually optimal for the racing environment.

\subsection{Learning-Based Model for Trajectory Planning}\label{Sec:gpr_plan}
Since modeling errors can lead to suboptimal and infeasible planned trajectories, we augment the time-optimal trajectory optimization problem with a data-driven modeling error compensation term using GPR. In particular, in the modeling constraint \eqref{Equ:TimeObj_constr2}, we replace $f(\cdot, \cdot)$ with a compensated dynamic model for planning, $f_{{plan}}(\cdot, \cdot)$:
\begin{equation} \label{equ:plan_gpr}
     f_{{plan}}(\mb{x}_k,\mb{u}_k) = f(\mb{x}_k,\mb{u}_k) + \boldsymbol{\mu}^{plan}(\mb{z}_{plan}(\hat{\mb{x}}_k, \hat{\mb{u}}_k)),
\end{equation}
where $\boldsymbol{\mu}^{plan}$ is the mean of the error-compensating GPR model:
\begin{equation}\label{equ:plan_gpr_model}
\begin{aligned}
\mb{g}^{plan}(\mb{z})\sim 
\mathcal{N}(\boldsymbol{\mu}^{plan}(\mb{z}), \mb{\Sigma}^{plan}(\mb{z})).
\end{aligned}
\end{equation}
Similar to \cite{hewing2018cautious}, we restrict $\boldsymbol{\mu}^{plan}$ to only compensate a select number of states, namely $\dot{V_y}$ and $\ddot{\psi}$ since they have the most significant nominal modeling error; therefore, $\boldsymbol{\mu}^{plan}=\left[0;\mu^{plan}_{\dot{V_y}};\mu^{plan}_{\ddot{\psi}};0;0;0\right]^\intercal$.

The feature mapping function $\mb{z}_{plan}(\cdot,\cdot): \mathbb{R}^{n_s}, \mathbb{R}^{n_u}\rightarrow \mathbb{R}^{n_f} $ maps the state vector of dimension $n_s$ and the control action vector of dimension $n_u$ to the feature vector. To find an informative set of features, we design several candidate features and determine which features to include by analyzing the correlation coefficients between state errors and candidate features, and select the candidates that have the highest correlation with the state error as our features, as described in Section \ref{subsec:state_feature_select}. It is worth noting that the compensation term is defined with respect to a constant nominal state vector $\hat{\mb{x}}_k$ and a nominal control action vector $\hat{\mb{u}}_k$ rather than the actual $\mb{x}_k$ and $\mb{u}_k$, so that the GPR model is not directly embedded into the optimization problem. Instead, we just use the GPR model to predict a constant compensation term based on the nominal value. In planning, we set the nominal state and action vectors as the solution from the last iteration of time-optimal optimization~\cite{ourpriorwork}. 

Using \eqref{equ:plan_gpr}, the GPR model is combined with the nominal model $f(\cdot, \cdot)$ to compensate for modeling error with respect to the observed dataset. In particular, the dataset $\mathcal{D}^{plan}$ for the GPR model \eqref{equ:plan_gpr_model} consists of the observed states and actions, $\mb{x}_k$ and $\mb{u}_k$, and the output data point, $\mb{y}^{plan}_k$, defined as
\begin{equation}\label{equ:dtp_plan}
    \mb{y}^{plan}_k = \dot{\mb{x}}_k - f(\mb{x}_k,\mb{u}_k),
\end{equation}
where $\dot{\mb{x}}_k$ is the derivative of the observed state $\mb{x}_k$. 

\subsection{MPC Tracking Controller}
\label{Sec:Controller}
A reference trajectory is provided by the time-optimal planner (Section \ref{Sec:TrajPlan}) with planning modeling error compensation (Section \ref{Sec:gpr_plan}).
We now introduce the MPC controller that tracks this reference trajectory.

Following our prior work~\cite{ourpriorwork}, we adopt a racing MPC with coupled longitudinal and lateral vehicle models that aims to reduce the tracking error between the actual vehicle state and the reference trajectory. The tracking objective and constraints are as follows:
\begin{subequations}\label{Equ:MPCCons}
\begin{align} 
\min_{\{\mb{x}_k\}_{k=1}^{N_p},\{\mb{u}_{k}\}_{k=1}^{N_{p-1}}} &
\sum_{k=1}^{N_{p}} \|\mb{x}_{k}-\mb{x}_{k}^{\mathrm{ref}}\|^{2}_{w_\mb{x}}+w_{\dot{\delta}} \Delta{\delta}_{k}^{2} \label{Eqn:MPCObj} \\
\mathrm{s.t.} \quad &\mb{x}_{k+1}=A_{k} \mb{x}_{k}+B_{k} \mb{u}_{k}+D_{k} \label{Eqn:MPCConstr1}\\
        & w_{r,k} \leq e_{y,k} \leq w_{l,k} \label{Eqn:MPCConstr2}\\
& u_{\min,k} \leq u_k \leq u_{\max,k}, \label{Eqn:MPCConstr3}
\end{align}
\end{subequations}
where $N_p$ is the prediction horizon of MPC. The objective function \eqref{Eqn:MPCObj} reduces the error between each of the states, with a weight corresponding to each element of the weight vector $w_\mb{x}\in \mathbb{R}^{n_s}$ with the dimension of $n_s$ same as the state $\mb{x}_k$, and regularization with weight $w_{\dot{\delta}}$ on the finite difference of steering angle to reduce rapid changes in steering angle. The term $\Delta{\delta}_k$ is defined as $\Delta{\delta}_k=\delta_k - \delta_{k-1}$. Constraints \eqref{Eqn:MPCConstr2} and \eqref{Eqn:MPCConstr3} are analogous to \eqref{Equ:TimeObj_constr3} and \eqref{Equ:TimeObj_constr4}, respectively. Unlike the nonlinear model employed by the trajectory planner, the MPC replaces constraint \eqref{Equ:TimeObj_constr1} with a linearization of the model \eqref{Eqn:MPCConstr1} to ensure the problem can be solved in real-time as a quadratic program. 
The linear model is found by linearizing the model around a heuristic nominal trajectory, $\bar{\mb{x}}_k$, as in~\cite{ourpriorwork}.

\subsection{Learning-Based Model for Control}\label{subsec:MPC_learning}
The MPC in Section \ref{Sec:Controller} finds the best control actions by optimizing the tracking problem with respect to the model constraint \eqref{Eqn:MPCConstr1}. However, the control actions may not actually be optimal if \eqref{Eqn:MPCConstr1} does not accurately model the system. We therefore introduce an error compensation term in the MPC vehicle model constraint \eqref{Eqn:MPCConstr1} as:
\begin{equation}
    \mb{x}_{k+1}=\mb{A}_{k} \mb{x}_{k}+\mb{B}_{k} \mb{u}_{k}+\mb{D}_{k} + \boldsymbol{\mu}^{MPC}(\mb{z}_k(\bar{\mb{x}}_k, \bar{\mb{u}}_k)), 
\end{equation}
where $\boldsymbol{\mu}^{MPC}$ is the mean of the GPR error model 
\begin{equation}
\begin{aligned}
\mb{g}^{MPC}(\mb{z})\sim 
\mathcal{N}\left(\boldsymbol{\mu}^{MPC}(\mb{z}), \mb{\Sigma}^{MPC}(\mb{z})\right).
\end{aligned}
\end{equation}
Similar to the GPR compensation function in planning, we restrict $\boldsymbol{\mu}^{plan}$ to only compensate $\dot{V_y}$ and $\ddot{\psi}$; therefore, $\boldsymbol{\mu}^{MPC} =\left[0;\mu^{plan}_{V_y};\mu^{MPC}_{\dot{\psi}};0;0;0\right]$. Also, we define a similar feature mapping function $\mb{z}_{MPC}(\cdot,\cdot): \mathbb{R}^{n_s}\times \mathbb{R}^{n_u}\rightarrow \mathbb{R}^{n_f}$. Similar to the case in planning, the compensation term is defined with respect to nominal state and action vectors. Here the nominal state and action vectors are chosen as the ones used for model linearization in the last subsection. 

To fit the GPR compensation function, we constructed a dataset $\mathcal{D}^{MPC}$ from the trajectories collected online. For each observed state $\mb{x}_{k+1}$, the output data point $\mb{y}^{MPC}_k$ in the dataset $\mathcal{D}^{MPC}$ is calculated as
\begin{equation}\label{equ:dtp_mpc}
\mb{y}^{MPC}_k = \mb{x}_{k+1} - (\mb{A}_{k} \mb{x}_{k}+\mb{B}_{k} \mb{u}_{k}+\mb{D}_{k}).
\end{equation}

\subsection{Iterative GPR Model Update Framework}\label{subsec:framework}
\begin{figure}[t]
    \centering
    \includegraphics[width=0.85\columnwidth]{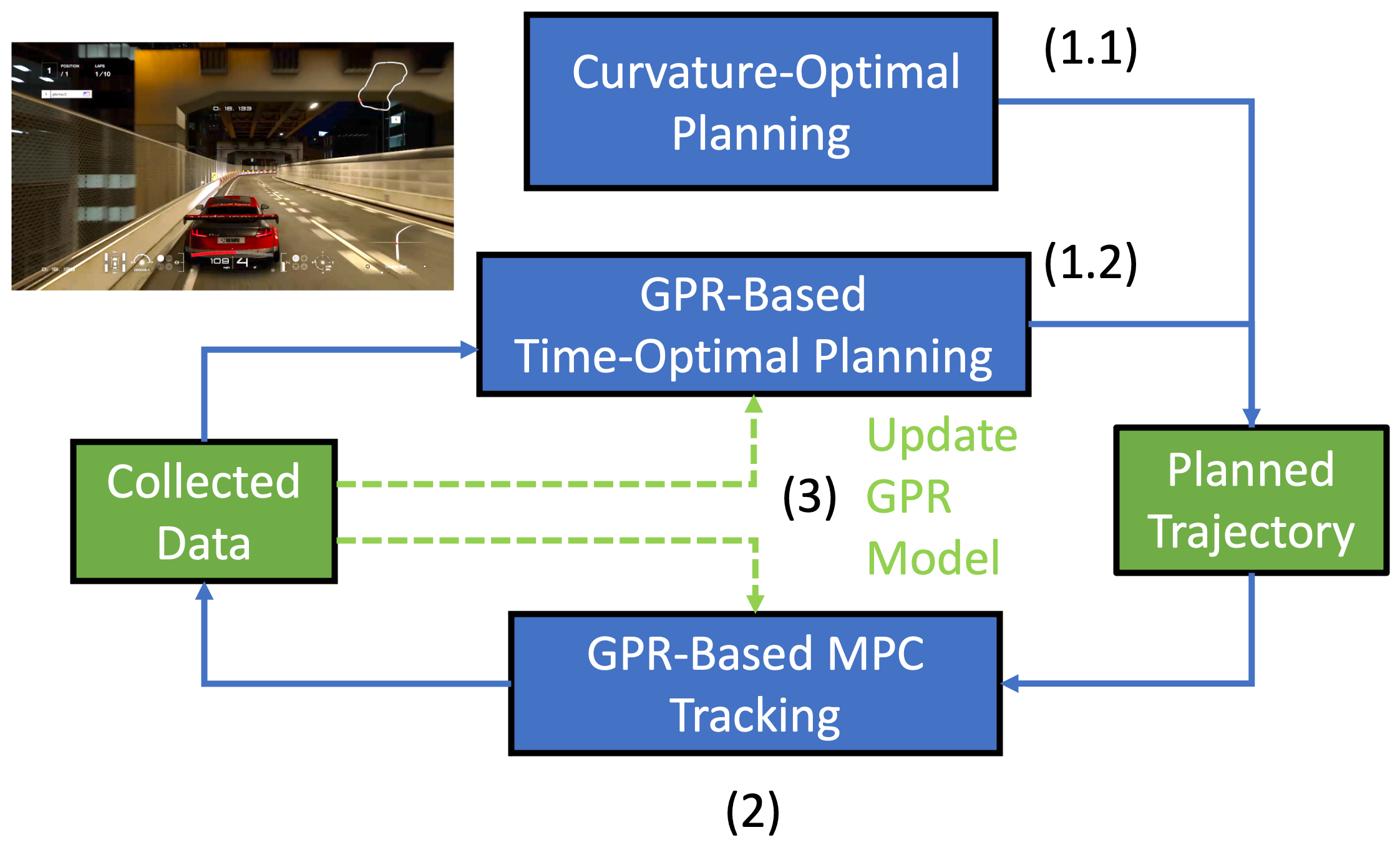}
    \caption{Iterative, double-GPR planning and control framework. \textbf{(1.1)}: At the $0^\text{th}$ iteration, curvature-optimal trajectory planning~\cite{ourpriorwork} without GPR error compensation; \textbf{(1.2)}: After $1^\text{st}$ iteration, time-optimal trajectory planning with GPR error compensation; \textbf{(2)}: Trajectory tracking GPR error compensation in MPC;\;\textbf{(3)}: Update of GPR error compensation models with tracking data.}\label{Fig:framework}
\end{figure}

So far we have detailed the trajectory planner and controller, each with corresponding error compensation functions using GPR. We now describe an iterative data collection scheme that ensures the GPR compensation functions are representative of the critical data regions for racing. The iterative framework is shown in Fig. \ref{Fig:framework}.
First in iteration 0, without exploiting GPR-based modeling error compensation, we plan a curvature-optimal trajectory~\cite{ourpriorwork} (\textbf{Step (1.1)}) and use the nominal MPC controller (Section \ref{Sec:Controller}) to track that trajectory (\textbf{Step (2)}) to collect data. After obtaining the initial dataset in the 0th iteration, we use it to build the initial GPR error compensation models (\textbf{Step (3)}) and apply them in the time-optimal trajectory planning (\textbf{Step (1.2)}) and subsequently MPC tracking (\textbf{Step (2)}). The closed-loop trajectories are added to the datasets for GPR. We use the augmented dataset to update our GPR model (\textbf{Step (3)}) and iteratively repeat \textbf{Step (1.2)}, \textbf{(2)}, and \textbf{(3)}.  Namely, The dataset $\mathcal{D}$ after $i^\text{th}$ iteration is updated as
\begin{equation}\label{eq:dataset}
    \begin{split}
     \mb{Y}^i &= [\mb{Y}_0; \mb{Y}_1; \mb{Y}_2;\dots; \mb{Y}_i] \\
     \mb{Z}^i &= [\mb{Z}_0; \mb{Z}_1; \mb{Z}_2;\dots; \mb{Z}_i] \\
    \end{split}
\end{equation}
where $\mb{Y}^i$ and $\mb{Z}^i$ are the updated output and input dataset after the $i^\text{th}$ iteration, $\mb{Y}_i$ and $\mb{Z}_i$ are the new output and input data collected in the $i^\text{th}$ iteration.  

This iterative framework allows us to continuously update our GPR models for both planning and control using the most recent and historical data; this leads to improved modeling accuracy in both planning and control, and therefore improves the overall performance of the planning-control framework. It is worth noting that we choose to track the curvature-optimal trajectory (as opposed to a time-optimal trajectory) in iteration 0 because the data collected by tracking this alternative reference enriches data diversity, which was shown in our experiments to improve the prediction performance of the models. To ensure accurate prediction over a long racing track, we maintain a dataset of over 10k data points for GPR fitting. Managing such a large dataset for real-time updates poses a challenge to the hardware. Instead, we update the dataset after each lap offline in our experiments. After updating the dataset, we randomly select a subset of data with a maximum size of 4000 samples to tune the hyperparameters as described in~\eqref{para_optm_func}.

\section{Experiment}\label{Sec:Experiment}
In this section, we test our \textit{iterative, double-GPR} planning and control framework with real-time experiments detailed in the following subsections. We compare our proposed \textit{iterative, double}-GPR framework, denoted as Double-GPR in the following subsections, to three baselines: 1) nominal-model planning and control using the methods in Section \ref{Sec:TrajPlan} and \ref{Sec:Controller} without any error compensation (Non-GPR), 2) learning-based planning with nominal-model control by introducing only the compensation function in \ref{Sec:gpr_plan} to the first baseline (GPR-Plan), and 3) nominal-model planning with learning-based control by introducing only the compensation function in \ref{subsec:MPC_learning} to the first baseline (GPR-MPC). Notably, our proposed framework outperforms these baselines algorithms, and the iterative framework in Section (\ref{subsec:framework}) improves the performance for the GPR-Plan, GPR-MPC, and Double-GPR methods. 

\subsection{Experimental Implementation}
\begin{figure}[t]
    \centering
    \includegraphics[width=0.9\columnwidth]{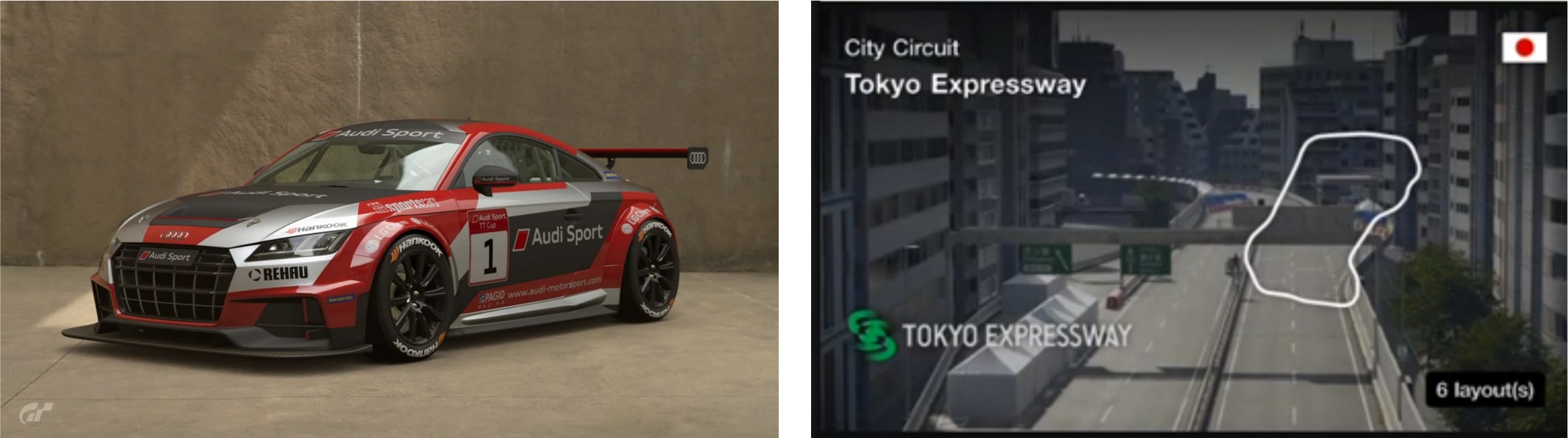}
    \caption{Racing vehicle: Audi TT Cup (left) and racing track: Tokyo Expressway Central Outer Loop (right).}
    \label{Fig:car_track}
\end{figure}

The closed-loop experiments were conducted in the high-fidelity racing platform Gran Turismo Sport (GTS)~\cite{GTS_web}, which is a world-leading racing simulator and is famous for its realistic modeling of racing vehicles.  
We utilize an \textit{Audi TT Cup} in GTS on the \textit{Tokyo Expressway Central Outer Loop} track, as depicted in Fig. \ref{Fig:car_track}.
The experimental computer is Alienware-R13 with Intel i9-12900 CPU and Nvidia 3090 GPU with Ubuntu 20.04 and Python 3.8.
The Sony Play Station 4 simulator hardware is connected to the computer  via a wired router.
We use cvxpy~\cite{diamond2016cvxpy} and qpsolver~\cite{domahidi2013ecos} to solve the QP problem in MPC and 
Cadasi~\cite{Andersson2019, wachter2006implementation} to solve the nonlinear programming problems in time-optimal planning.
We identified the vehicle parameters of the nominal vehicle model based on data collected with a nominal controller. The identified parameters are summarized in Table~\ref{tab:Parameters}.

\renewcommand\arraystretch{1.2} 
\begin{table}[b]
    \centering
    \caption{Parameters of Audi TT Cup in GTS}
    \begin{tabular}{lc}
    \hline \hline
    Parameter & Value \\
    \hline
    Total mass $m$ &  1161.25 $kg$  \\
    Length from CoG to front wheel $l_f$ & 1.0234 $m$ \\
    Length from CoG to rear wheel $l_r$ & 1.4826 $m$ \\
    Width of chassis &1.983 $m$ \\
    Height of CoG $h_c$ &  0.5136 $m$ \\
    Friction ratio $\mu$ & 1.5\\
    Wind drag coefficient $C_{xw}$ & 0.1412 $kg/m$ \\
    Moment of inertia $I_{zz}$ & 2106.9543 $Nm$ \\
    \hline \hline
    \end{tabular}
    \label{tab:Parameters}
\end{table}

To make real-time GPR estimation feasible, we exploit the use of GPU and adopt an efficient and general approximation of GPR based on Blackbox Matrix-Matrix multiplication (BBMM) \cite{gardner2018gpytorch}, which supports over 10,000 data points in the GPR dataset while keeping almost the same prediction accuracy with original GPR; we find it outperforms other approximation methods that significantly reduced our modeling accuracy and only supported significantly smaller datasets \cite{hewing2018cautious}. Our GPR implementation can then be easily embedded in the MPC controller operating at a 20 Hz control frequency.

For Double-GPR, we observed that the GPR-based MPC controller was able to track the planned trajectories with small tracking errors even in the first several iterations. The collected data then concentrated around the planned trajectories and, thus, were not sufficiently diverse to fit a GPR model that was globally accurate. As a result, we observed that the GPR-based time-optimal planning may lead to infeasible planned trajectories if they deviated too much from the data support. To address this problem, in each iteration, we warm-started time-optimal planning with the planned trajectory from the last iteration and only applied a \emph{single-step} update in time-optimal planning. We found this strategy could effectively stabilize Double-GPR and lead to a converged closed-loop lap time. 

\subsection{Compensation Feature Selection}\label{subsec:state_feature_select}
To find out informative features that are useful for error prediction, we first select a set of candidate features collected in the vector $\mb{z}^{cand} = [V_x, V_y, \dot{\psi}, \delta, a_p, \dot{\delta}, \dot{a}_p]$, where $ \dot{\delta}$ is the steering rate, $a_p$ is the pedal command, and $\dot{a}_p$ is the changing rate of the pedal command. Then we estimate the correlation between the state prediction errors and the candidate features to filter out uninformative features. The correlation is estimated empirically using a pre-collected dataset. The state errors of the models for planning and control are essentially the ones defined in~\eqref{equ:dtp_plan} and~\eqref{equ:dtp_mpc}, i.e., $\mb{e}^{plan} = \mb{y}^{plan}$, $\mb{e}^{MPC} = \mb{y}^{MPC}$. 
In addition to this, we also refine feature selection according to the results of experiments.
Eventually, we observe that the errors of the state in both models are most significantly impacted by $V_y$, $\dot{\psi}$, and $\delta$.
Therefore, in this work, we choose $\mb{z} =[V_y, \dot{\psi}, \delta ]$ as the feature vector.


\subsection{Planning and Control Result}
For each method, we implement the iterative framework detailed in Section \ref{subsec:framework}. For a fair comparison, we construct the initial dataset for all GPR-based methods using the same collected data in \textbf{Step (1.1)}. We then repeat the iterative framework in Fig. \ref{Fig:framework} for 10 iterations, alternating between time-optimal planning, control, and re-calculation of GPR compensation functions. It is worth noting that due to slight uncertainty in the timing of network communications with the  PS4, the GTS environment can be thought of as stochastic; this stochastic environment leads to slight variation in results from lap to lap (iteration to iteration), which we analyze below.

\begin{figure*}[t]
    \centering
    \includegraphics[width=0.28\textwidth]{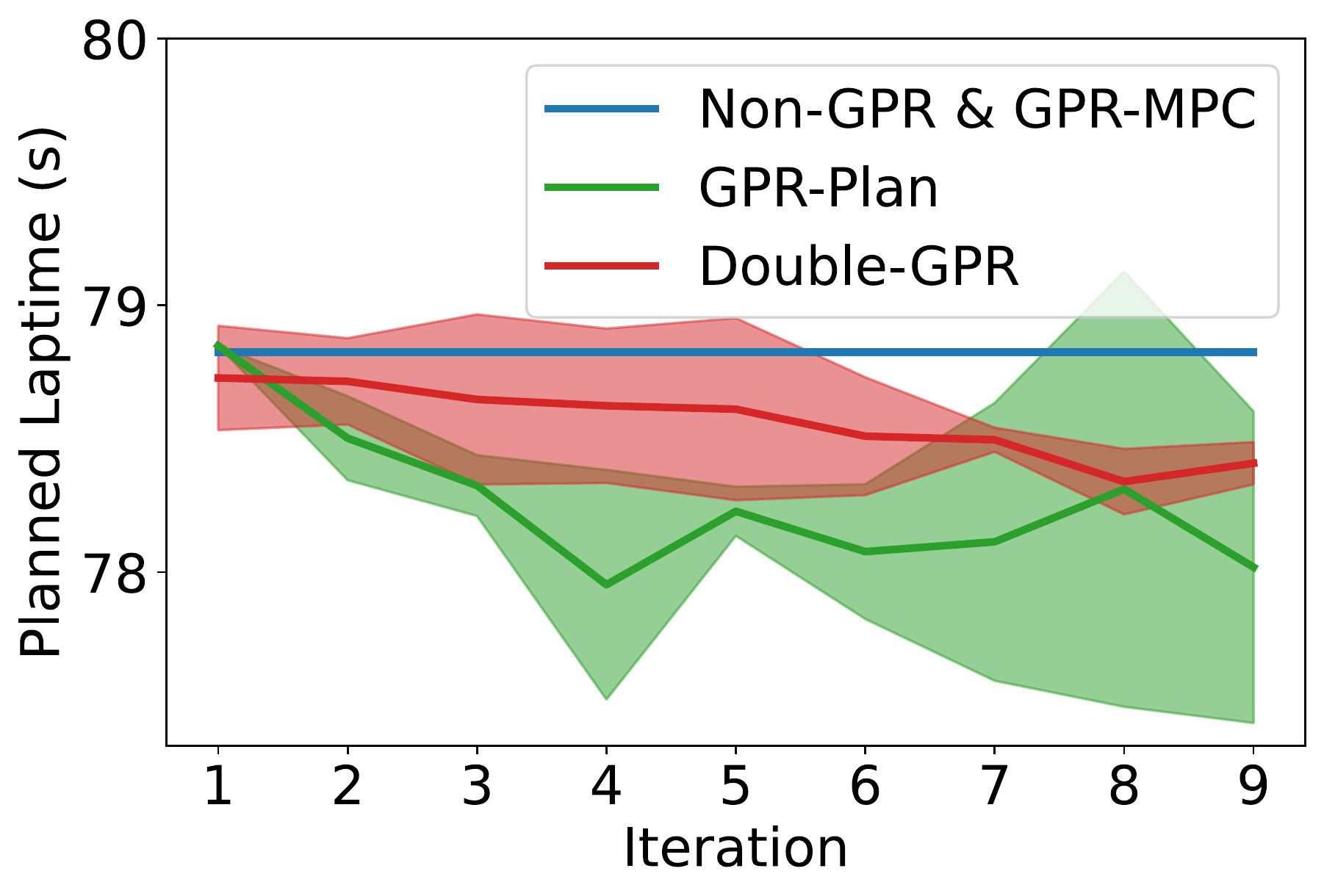}
    \hfil
    \includegraphics[width=0.28\textwidth]{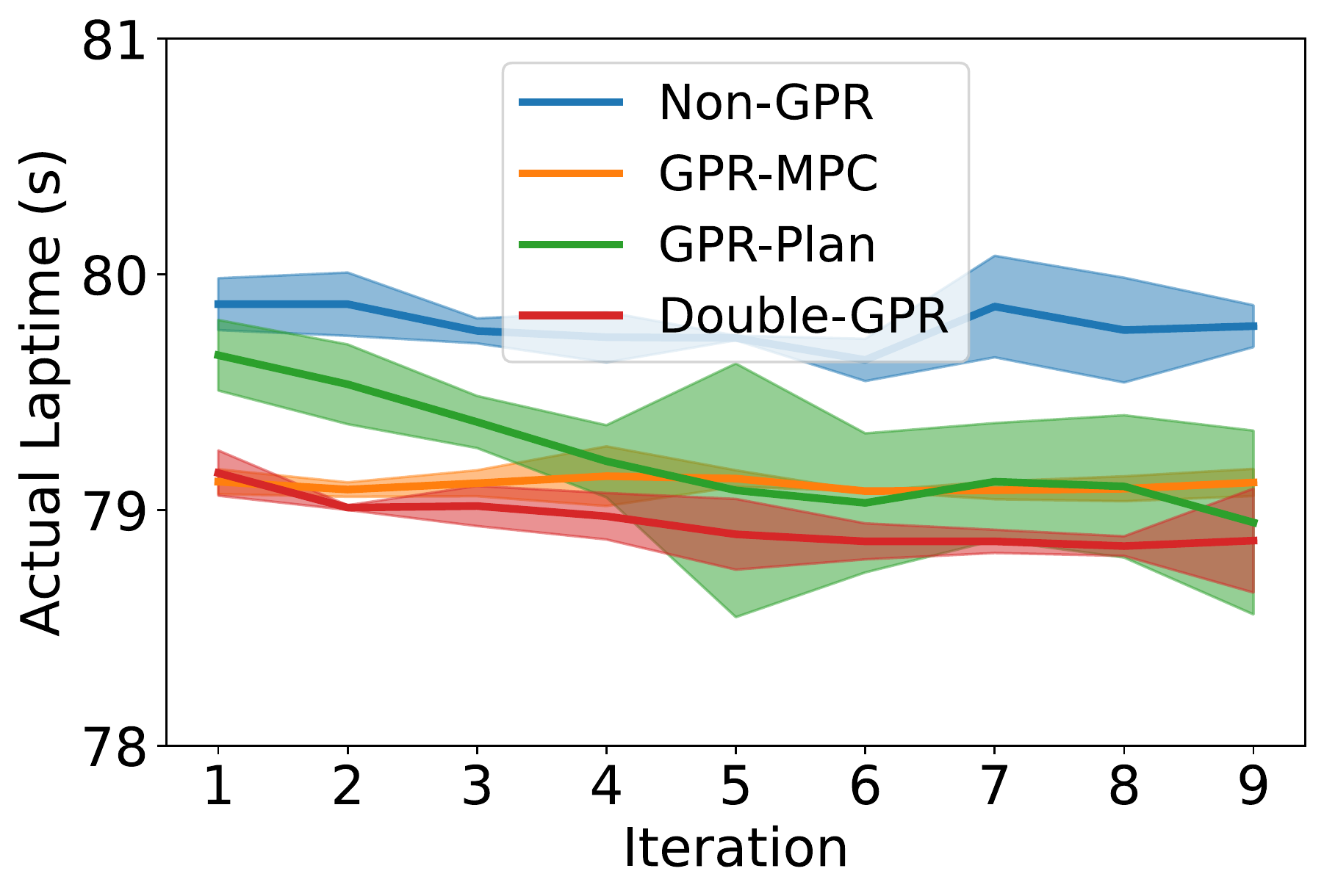}
    \hfil
    \includegraphics[width=0.28\textwidth]{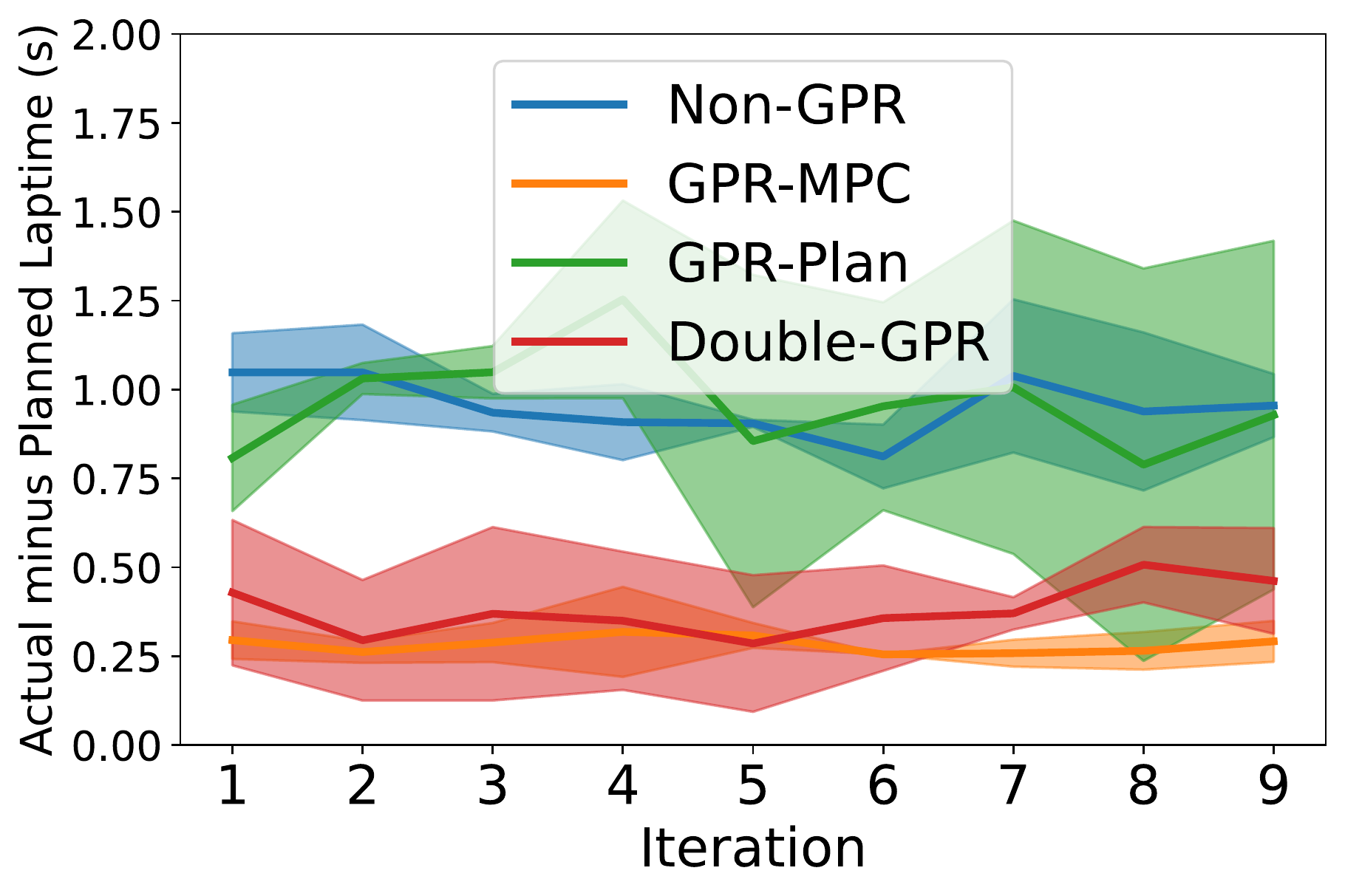}
    \caption{
    Lap time of planned trajectories (left), actual control experiments  in GTS (middle) and the difference in lap time between planned reference and actual experiment (right). Learning-based framework in Fig. \ref{Fig:framework} is repeated for three trials; solid line indicates mean  and error range represents standard deviation over the three trials.}
    \label{Fig:lap_time_result}
\end{figure*}

We first present the planned and actual lap time for each of the methods with respect to iterations in Fig.~\ref{Fig:lap_time_result}. The \textit{Planned Laptime} for each iteration represents the lap time expected from the planning module in \textbf{Step (1.2)}. Since the GPR-MPC and Non-GPR methods do not implement error compensation for the planning module, the planned lap time is identical across all iterations. 
The \textit{Actual Laptime} for each iteration represents the closed-loop lap time obtained by tracking the planned trajectory with the control module in \textbf{Step (2)}. 
Our proposed Double-GPR method shows the most consistent improvement in both the planned and actual laptime as the dataset expands over each iteration. 
More importantly, the time gap between the planned and actual laptime consistently decreases when Double-GPR is used. It is crucially important for pushing the race car toward its performance upper bound. 

\begin{figure}[t]
\centering 
    \begin{subfigure}{\columnwidth}
    \centering
    \includegraphics[width=\columnwidth]{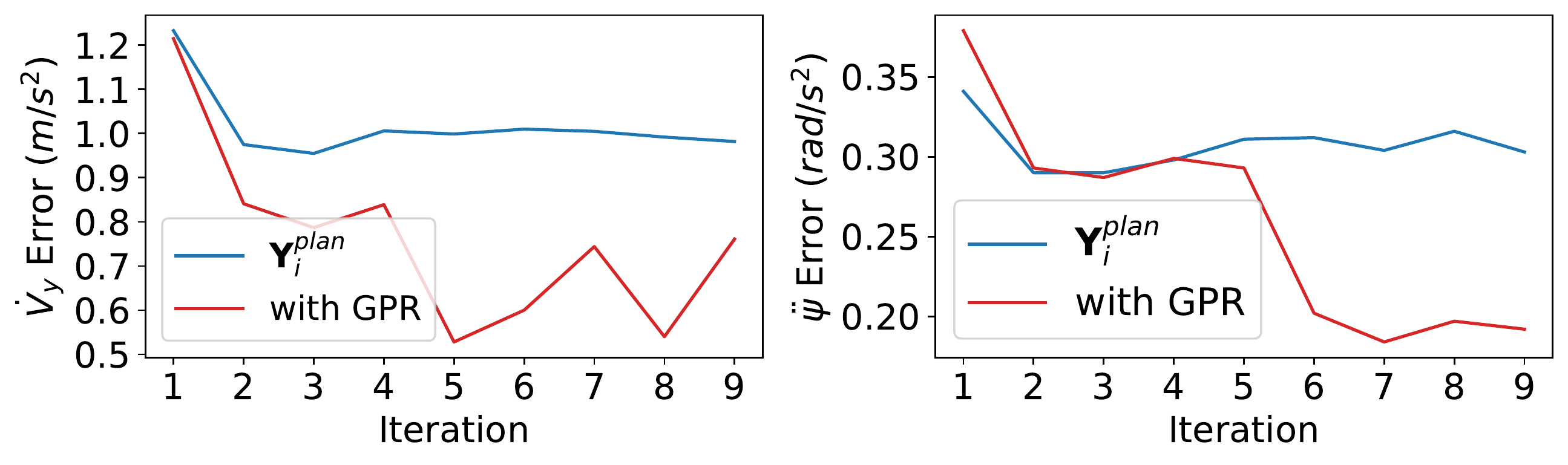}
    \caption{Planning Model Error}
    \end{subfigure} %
    \begin{subfigure}{\columnwidth}
    \centering
    \includegraphics[width=\columnwidth]{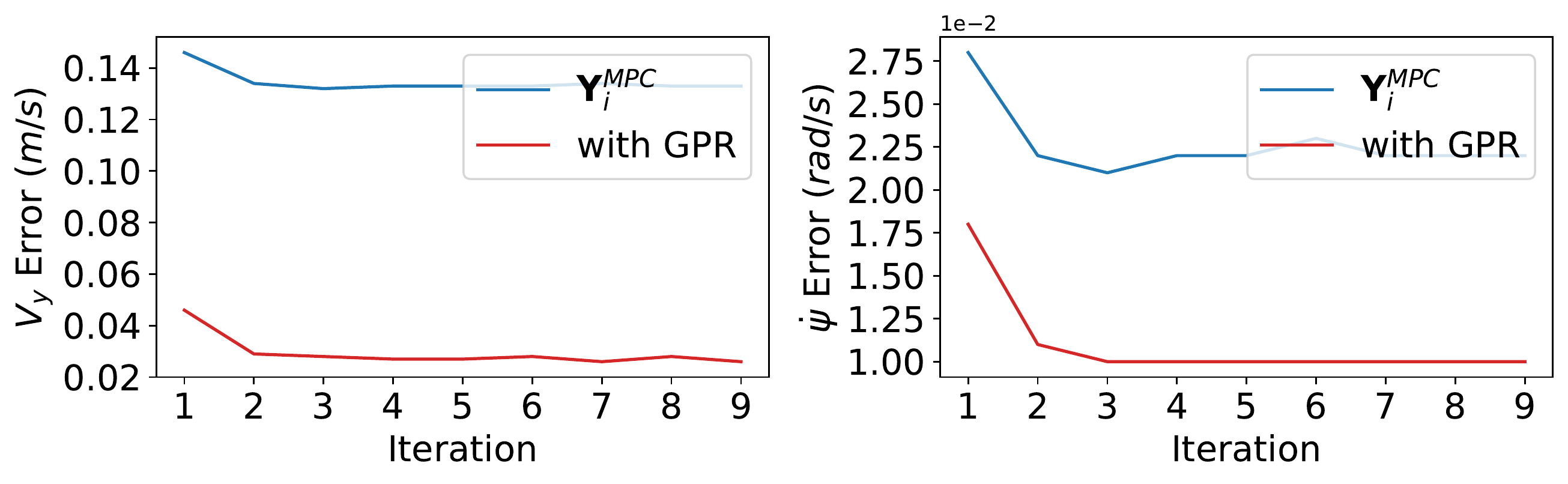}
    \caption{MPC Model Error}
    \end{subfigure}
    \caption{Average model error of Double-GPR with and without GPR compensation in each iteration.}\label{Fig:model_error}
\end{figure}

We also examined the modeling errors over iterations of Double-GPR. The results are plotted in Figure \ref{Fig:model_error}. Here, $\mb{Y}^{Plan}_i$ and $\mb{Y}^{MPC}_i$ represent the errors of the nominal models in the data collected from $i^\mathrm{th}$ iteration as defined in \eqref{eq:dataset}. We compare the errors of the nominal models with the modeling errors after compensated by the GPR functions at each iteration. As shown in the figure, the GPR compensation functions indeed reduce the modeling errors for both planning and control, and the models become more accurate over the iterations.

\renewcommand\arraystretch{1.5}
\begin{table*}[th]
\caption{{Lap time and tracking result}} \label{tab:combine}
    \centering
\begin{threeparttable}
    \begin{tabular}{ccccccccc}
    \hline \hline
        & \makecell[c]{Planning \\  lap time($s$)} & \makecell[c]{Closed-loop \\  lap time($s$)\tnote{1}}  & \makecell[c]{Lap time \\  difference($s$)\tnote{1}}  & \makecell[c]{ Best lap time\\  difference($s$) }   & \makecell[c]{$e_y$($m$)\\  Tracking Error\tnote{2} } & \makecell[c]{$e_{V_x}$($m/s$) \\ Tracking Error\tnote{2} }  & \makecell[c]{$e_\psi$ ($rad$)\\ Tracking Error\tnote{2}  }     \\
    \hline 
    \textbf{Non-GPR} & \color{black}{78.825}  & \color{black}{79.779$\pm$0.134} & \color{black}{0.954$\pm$0.134} & \color{black}{0.755} & \color{black}{0.253$\pm$0.011}  & \color{black}{1.211$\pm$0.079} & \color{black}{0.644$\pm$0.027} \\
    
    \textbf{GPR-MPC} &  \color{black}{78.825} & \color{black}{79.107$\pm$0.054} & \color{black}{0.282$\pm$0.054} & \color{black}{0.205} & \color{black}{0.229$\pm$0.015} & \color{black}{0.808$\pm$0.024} & \color{black}{0.695$\pm$0.021} \\
    
    \textbf{GPR-Plan} & \color{black}{78.264$\pm$0.439} & \color{black}{79.228$\pm$0.336} & \color{black}{0.964$\pm$0.331} & \color{black}{0.262} & \color{black}{0.278$\pm$0.044} & \color{black}{1.300$\pm$0.259} & \color{black}{0.755$\pm$0.112} \\
    
    \textbf{Double-GPR} & \color{black}{78.564$\pm$0.225} & \color{black}{78.945$\pm$0.134} & \color{black}{0.380$\pm$0.159} & \color{black}{0.066} & \color{black}{0.241$\pm$0.035} & \color{black}{0.851$\pm$0.077} & \color{black}{0.719$\pm$0.061} \\
    \hline \hline
\end{tabular}

\begin{tablenotes}
    \item[1] The closed-loop lap time and lap time difference are presented in the format of $\mathrm{mean} \pm \mathrm{std}$. The statistics are computed over 10 trials.
    \item[2] Lateral distance error $e_y$, longitudinal velocity error $e_{V_x}$, and heading error $e_\psi$ are presented in the format of $\mathrm{mean} \pm \mathrm{std}$. The statistics are computed over 10 trials.
\end{tablenotes}
    
\end{threeparttable}
\end{table*}

We present the average results over all iterations in Table~\ref{tab:combine}. In comparison to the Non-GPR, incorporating GPR compensation in either planning or control can substantially reduce the closed-loop lap time. Specifically, GPR-MPC and Double-GPR can significantly narrow the gap between the planned and actual lap times. Among all methods, Double-GPR achieves a minimum time gap of $0.066s$ and achieved the shortest closed-loop lap time of $78.945s$, which supports our argument that incorporating modeling compensation for both planning and control is important for achieving optimal racing performance.
Table \ref{tab:combine} additionally lists the mean and standard deviation of the tracking errors of the MPC. By compensating for modeling errors in the MPC, Double-GPR reduces the tracking error compared to GPR-Plan, indicating a more stable tracking performance across all iterations.

\section{Conclusion}\label{Sec:conclusion}

In this paper, we propose an \textit{iterative, double-GPR} modeling error compensation framework; our framework compensates for modeling errors in \textit{both} offline long-horizon time-optimal trajectory planning \textit{and} short-term MPC tracking. The framework enables iterative improvement of modeling accuracy and racing performance through data collected online. We test our proposed framework for autonomous racing in the high fidelity racing simulation game Gran Turismo Sport. Through comparison of our double-GPR method to methods including compensation for just the planning module or the controller module, we find that having double-GPR compensation functions for both planning and control is critical for accurate modeling and achieving optimal laptime.


\section*{Acknowledgment}
This work was supported by Sony AI. We would like to thank Kenta Kawamoto from Sony AI for his kind help and fruitful discussions. We are also very grateful to Polyphony Digital Inc. for enabling this research and providing the GT Sport framework. This material is based upon work supported by the National Science Foundation Graduate Research Fellowship Program under Grant No. DGE 1752814. Any opinions, findings, conclusions, or recommendations expressed in this material are those of the authors and do not necessarily reflect the views of the National Science Foundation.

\bibliographystyle{IEEEtran}
\bibliography{references}

\end{document}